\documentclass[journal]{IEEEtran}
\usepackage{bm}
\usepackage{amsmath}
\usepackage{amssymb}
\usepackage{amsthm}
\usepackage{mathrsfs}
\usepackage{enumerate}
\usepackage{multirow}
\usepackage{color}
\usepackage{threeparttable}
\usepackage{subfigure}

\usepackage[square, comma, sort&compress, numbers]{natbib}
\usepackage[pagebackref=false,breaklinks=true,letterpaper=true,colorlinks,bookmarks=false]{hyperref}

\usepackage{times}
\usepackage{breakurl}
\usepackage{array}
\usepackage{verbatim}
\usepackage{algorithm}
\usepackage{bbding}
\usepackage{algpseudocode}
\usepackage{lettrine}
\usepackage{hyperref}
\ifCLASSINFOpdf
  \usepackage[pdftex]{graphicx}
  \usepackage{graphics}
  \usepackage{graphicx}
  \usepackage{epsfig}
  \usepackage{epstopdf}
  \graphicspath{{./fig/}}
  \DeclareGraphicsExtensions{.pdf,.jpeg,.png,.eps}
\else
  \usepackage[dvips]{graphicx}
  \graphicspath{{./fig/}}
  \DeclareGraphicsExtensions{.eps}
\fi

\ifCLASSINFOpdf
\else
\fi

\begin{document}
\title{Triplet Contrastive Representation Learning for Unsupervised Vehicle Re-identification}

\author{Fei Shen,
        Xiaoyu Du,
        Liyan Zhang,
        Xiangbo Shu,
        and Jinhui Tang
    % <-this % stops a space
\thanks{Fei Shen, Xiaoyu Du, Xiangbo Shu, and Jinhui Tang are with the School of Computer Science and Engineering, Nanjing University of Science and Technology, Nanjing, 210094, China. e-mail: {feishen@njust.edu.cn; duxy@njust.edu.cn; shuxb@njust.edu.cn; jinhuitang@njust.edu.cn}.}
\thanks{Liyan Zhang is with the College of Computer Science and Technology,
Nanjing University of Aeronautics and Astronautics, Nanjing, 210016, China. e-mail: zhangliyan@nuaa.edu.cn. (Corresponding author: Liyan Zhang.)}}
% <-this % stops a space

\markboth{Journal of \LaTeX\ Class Files,~Vol.~14, No.~8, August~2015}%
{Shell \MakeLowercase{\textit{et al.}}: Bare Demo of IEEEtran.cls for IEEE Journals}

\maketitle


\begin{abstract}
Part feature learning is critical for fine-grained semantic understanding in vehicle re-identification.
However, existing approaches directly model part features and global features, which can easily lead to serious gradient vanishing issues due to their unequal feature information and unreliable pseudo-labels for unsupervised vehicle re-identification.
To address this problem, in this paper, we propose a simple Triplet Contrastive Representation Learning ~(TCRL) framework which leverages cluster features to bridge the part features and global features for unsupervised vehicle re-identification. Specifically, TCRL devises three memory banks to store the instance/cluster features and proposes a Proxy Contrastive Loss~(PCL) to make contrastive learning between
adjacent memory banks, thus presenting the associations between the part and global features as a transition of the part-cluster and cluster-global associations.
Since the cluster memory bank copes with all the vehicle features, it can summarize them into a discriminative feature representation.
To deeply exploit the instance/cluster information, TCRL proposes two additional loss functions. For the instance-level feature, a Hybrid Contrastive Loss (HCL) re-defines the sample correlations by approaching the positive instance features and pushing the all negative instance features away. For the cluster-level feature, a Weighted Regularization Cluster Contrastive Loss (WRCCL) refines the pseudo labels by penalizing the mislabeled images according to the instance similarity.
Extensive experiments show that TCRL outperforms many state-of-the-art unsupervised vehicle re-identification approaches.

\end{abstract}


% \fi
\IEEEpeerreviewmaketitle

\section{Introduction}

Vehicle re-identification ~\cite{shen2020net, liu2019group,li2022attribute,wang2022attentive,guo2019two} aims to search for the querying vehicle from non-overlapping cameras. It has received wide-spread attention,  due to the rapidly growing requirements for traffic video surveillance.
The state-of-the-art approaches lie on the supervised learning \cite{luo2019strong,liu_gcn,vanet,vit-bot,emrn} and achieve excellent performance on the public vehicle datasets. However, these approaches require extremely time-consuming and labor-intensive data annotation which limits their use in real scenarios.
Therefore, the re-identification community~\cite{wang2022uncertainty,wang2021inter,li2021exploiting,zheng2020vehiclenet,lin2020aggregating,si2022hybrid,reid-survery,dynamicgcn,dgm,ye2018robust,tdsr,luo2021self} now pays wide attention to the unsupervised learning approaches to introduce the unlabeled data.

\begin{figure}[t]
	\centering
	\includegraphics[width=1.0\linewidth]{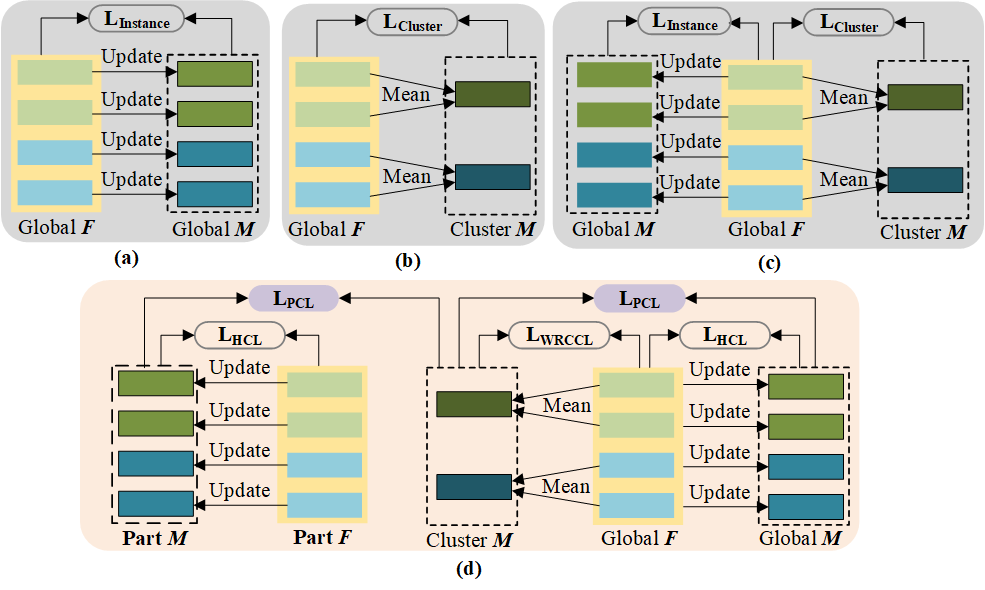}
		\vspace{-.3cm}
	\caption{Comparison of four types of memory-based contrastive learning methods.  Here, $\textbf{\emph{F}}$ and $\textbf{\emph{M}}$ respectively denote features and memory bank.
 (a) Instance contrastive learning. (b) Cluster contrastive learning. (c)  Dual contrastive learning. (d) Our proposed Triplet Contrastive Representation Learning (TCRL) establishes the connection between part features and global features via a proxy of a cluster memory bank. Simultaneously, it computes the loss and updates the features both at instance-level (\emph{including part and global}) and cluster-level.}
	\label{fig:example}
 		\vspace{-.3cm}
\end{figure}

%实例级对比学习
Contrastive learning is a major technique of unsupervised re-identification. They mostly utilize a memory bank~\cite{simclr, hhcl, cacl} to store the recent-step instance/cluster features for the next-step contrastive process.
The development of contrastive learning is divided into three stages from memory-based structure.
Fig.~\ref{fig:example} (a) demonstrates the instance-oriented approaches~\cite{moco,simclr,swav,byol,simsiam,bottom,ye2019unsupervised} that treat each image as a sole class and store all instance features in a global memory bank.
Fig.~\ref{fig:example} (b) demonstrates the cluster-oriented approaches~\cite{spcl,ca-ureid,ccl,ucf} that construct a cluster memory bank with average categorical features.
As the former neglects the categorical correlations among the images while the latter neglects the diversity of positive samples~(changes caused by perspectives, illuminations, scales, etc.), the dual contrastive \cite{mgce-hcl,hhcl,cacl} approaches shown in Fig. \ref{fig:example} (c) incorporates the global and cluster memory banks to deeply exploit the intra-class information.

\begin{figure*}[t]
	\centering
	\includegraphics[width=1.0\linewidth]{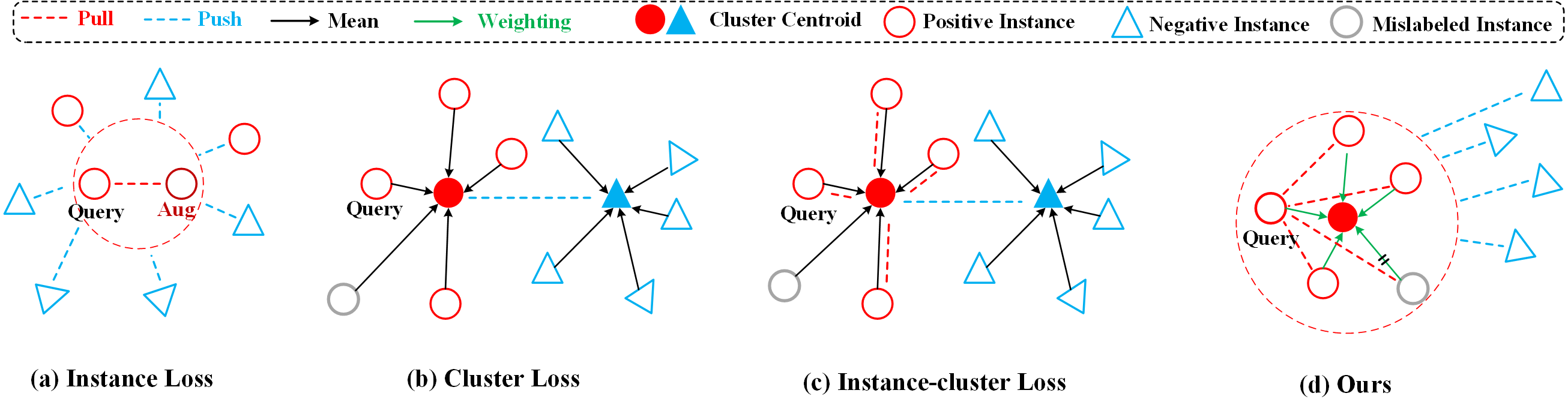}
		%\vspace{-.3cm}
	\caption{Illustration for different contrastive learning losses. Different colors and shapes denote different identities.  Ours contains proposed Hybrid Contrastive Loss~(HCL) and  Weighted Regularization Cluster Contrastive Loss~(WRCCL).  HCL closes the distance between query samples and instance-level features of positive samples, pushing all negative samples away.  WRCCL refines cluster-level sample correlation and penalizes the mislabeled images by weighting.}
	\label{fig:diff}
% 		\vspace{-.3cm}
\end{figure*}

Although the contrastive approaches achieve impressive performance, they neglect that the vehicle re-identification task is a fine-grained image retrieval task. Especially vehicles with the same model and color are hardly identified with global features. An intuitive solution is to introduce part and global features like supervised approaches~\cite{pcb,aaver,guo2019two,lou2019embedding,liu2019group,zhou2018vehicle,shen2020net,git,hsgm} at the same time. However, encoder directly models the part features and global features, which is prone to exhibit serious gradient collapse issues because the identity mapping makes them naturally fall into constants (\emph{trivial solutions}), just like ~\cite{simsiam, zbontar2021barlow, byol}.

To address the above issues, we propose a Triplet Contrastive Representation Learning ~(TCRL) framework
to establish the connection between part features and global features naturally via a proxy of a cluster memory bank.
As shown in Fig.~\ref{fig:example} (d), TCRL devises three
memory banks --- Part $M$, Cluster $M$, and Global $M$ --- to store the features of partial images, clustered centroids, and entire images, respectively. To model the part-cluster and cluster-global correlations across the memory banks, TCRL devises a Proxy Contrastive Loss~(PCL) that estimates the similarities with Kullback-Leibler Divergence and Euclidean Distance. As the cluster memory bank plays the intermediate role between the part and global memory bank and copes with all the features, it can summarize them into a final discriminative feature representation.

In addition, recent contrastive loss functions may mislead the learned instance correlations. As shown in Fig.~\ref{fig:diff}, we observe that a) the instance loss lacks the correlation between positive instances; b) the cluster loss concentrates on the cluster centroid only; and c) the instance-cluster loss neglects the influence of negative instances. %Therefore,
%we devote ourselves to devising more accurate correlation constraints for contrastive learning approaches. In addition,
Accordingly, we propose the Hybrid Contrastive Loss~(HCL) and Weighted Regularization Cluster Contrastive Loss~(WRCCL). As shown in Fig.~\ref{fig:diff} (d), HCL adequately exploits the negative information by directly comparing the query instance with all negative instances, and WRCCL penalizes the mislabeled instances via weighted correlations, respectively.

The main contributions of this paper are summarized as follows:
\begin{itemize}
\item
A simple Triplet Contrastive Representation Learning framework~(TCRL) is proposed to introduce the part features in learning vehicle representations. TCRL bridges the global and part features through three instance/cluster memory banks and proposes a Proxy Contrastive Loss~(PCL) to model the adjacent memory banks.

\item  We devise Hybrid Contrastive Loss~(HCL) and Weighted Regularization Cluster Contrastive Loss~(WRCCL) to re-define the instance/cluster correlations. HCL introduces the all individual negative instances into instance-level comparison. WRCCL weights the correlations to alleviate the impact of mislabeled images.

\item We conduct extensive experiments on three large-scale vehicle datasets to demonstrate that the proposed method is superior to the state-of-the-art unsupervised vehicle re-identification approaches.
\end{itemize}

\section{Related Work}
In this section, we illustrate the related works for vehicle re-identification. We first introduce the contrastive learning approaches in the instance, cluster, and dual learning perspectives. We then present the use of the part features in re-identification approaches.

\begin{figure*}[tp]
	\centering
	\includegraphics[width=1.0\linewidth]{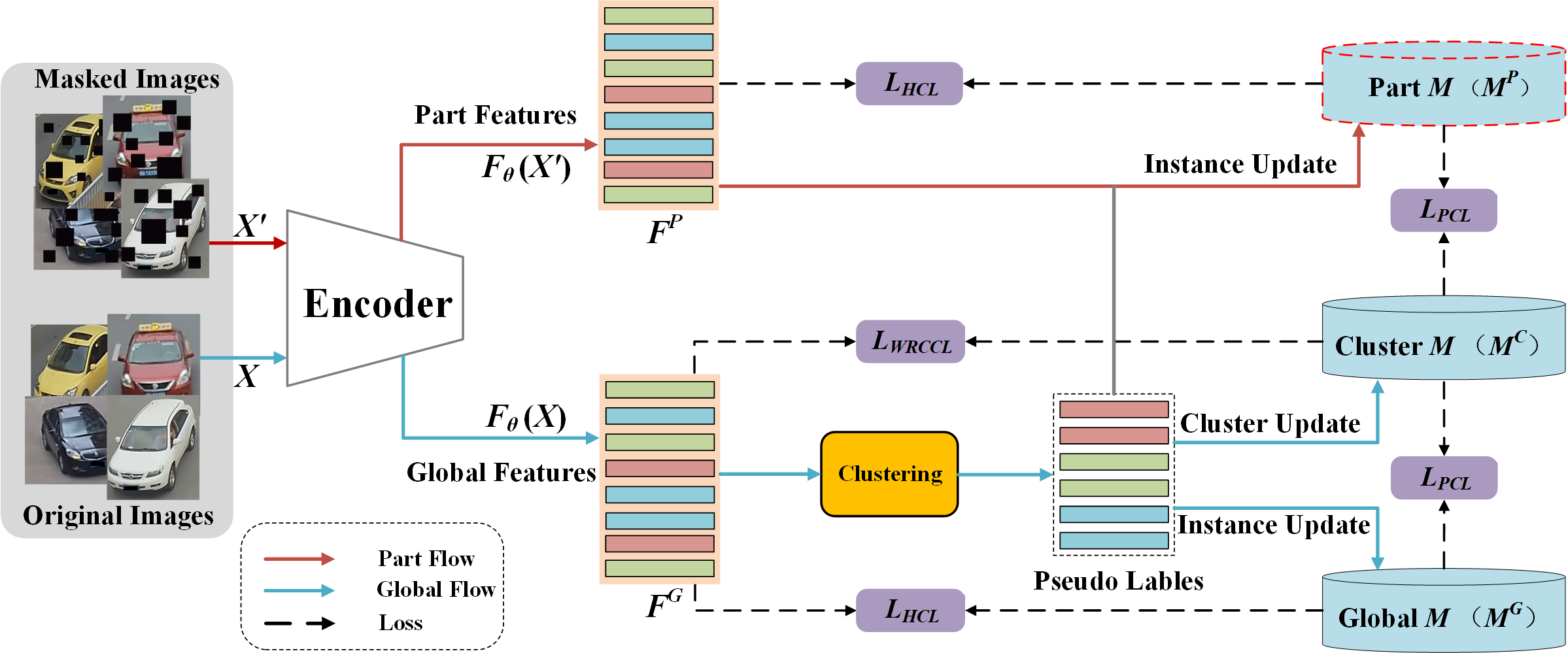}
		%\vspace{-.3cm}
	\caption{The framework of the proposed Triplet Contrastive  Representation Learning (TCRL), including part memory bank $M^P$, cluster memory bank $M^C$, and global memory bank $M^G$.
 Here, ${U^{P}}$and ${U^{G}}$  respectively part and global features.
 \textbf{Training:} Original images are first sampled in mini-batches to generate the corresponding masked images.
 Then, these two batches of images are fed to the encoder simultaneously to obtain global features and part features.
 Second,  a clustering algorithm is applied to cluster similar global features and assign pseudo labels to them.
 Third, the part feature of all samples, the average global feature of each cluster, and the global feature of all samples are stored in $M^P$, $M^C$, and $M^G$, respectively.
 Finally, the features of three memory banks are updated with momentum via our proposed three loss functions in TCRL.
 \textbf{Inference:}  Masked images and part features are only used for training and will be removed for a fair comparison. Thus, we extract features of test images through the encoder, and the cosine distance is applied as the similarity measurement.
	}
	\label{fig:framework}
% 		\vspace{-.4cm}
\end{figure*}

\subsection{Instance Contrastive Learning}
The instance contrastive learning methods \cite{moco,simclr,swav,byol,simsiam,bottom} regard each image as an individual class and consider two
augmented views of the same image as positive pairs and treat others in the same batch as negative pairs.
For example, momentum contrast (MoCo)~\cite{moco} transforms into a dictionary lookup task, using a contrastive loss to learn instance discriminative representations, treating each unlabeled example as a distinct class.
Simple framework for contrastive learning of visual representations (SimCLR)~\cite{simclr} regards samples in the current batch as the negative samples.
Similarly,
Bottom~\cite{bottom} treats each individual sample as a cluster and then progressively groups similar samples into a cluster, generating pseudo labels.
Though instance-level contrastive loss performs well in downstream tasks, it performs poorly on re-identification tasks that require correct measurement of inter-class differences on unsupervised target domains.
\subsection{Cluster Contrastive Learning}
The cluster contrastive learning methods~\cite{spcl,ca-ureid,ccl,ucf} are initialized with a cluster-level memory dictionary.
The clustering algorithms are used to generate corresponding pseudo labels in the above methods.
For example, cluster contrast learning~(CCL)~\cite{ccl} employs a unique cluster representation to describe each cluster, computing contrast loss at the cluster level.
Self-paced contrastive learning (SPCL)~\cite{spcl} proposes a novel self-paced contrastive learning framework that gradually creates a more reliable cluster to refine the memory dictionary features.
Uncertainty-aware clustering framework (UCF)~\cite{ucf} proposes a novel hierarchical clustering scheme to promote clustering quality and introduce an uncertainty-aware collaborative instance selection method.
\subsection{Dual Contrastive Learning}
Dual contrastive learning methods \cite{cacl, hhcl, dan, vapc} are typically initialized with a cluster-level memory dictionary and instance-level memory to distill the advantages from the two parts.
Cluster-guided asymmetric contrastive learning (CACL) \cite{cacl} designs an asymmetric contrastive learning framework to guide the siamese network effectively mine the invariance in feature representation.
 Hard-sample guided hybrid contrast learning (HHCL) \cite{hhcl} combines cluster centroid contrastive loss with hard instance contrastive loss for unsupervised person re-identification.
Besides, there are some others methods.
For example, the dual-branch adversarial network (DAN) \cite{dan} develops an image-to-image translation network without any annotation for unsupervised vehicle re-identification.
Viewpoint-aware progressive clustering (VAPC) \cite{vapc} divides the entire feature space into different subspaces and then performs a progressive clustering to mine the authentic relationship among samples.
However, the unsupervised vehicle re-identification approaches insufficiently model the part features thus impacting the final performance of unsupervised methods.

\subsection{Part Feature Learning}
Part feature learning methods usually divide feature maps into several parts and then individually pool each region, as done in~\cite{prn,hpgn,san,tamr,sff,li2022enhancing}. For example, stripe-based and attribute-aware network (SAN) \cite{san}  extracts the part features from the visual appearance of vehicles through a stripe-based branch and an attribute-aware branch. Hybrid pyramidal graph network (HPGN)~\cite{hpgn} explores the spatial significance of part features at multiple scales via spatial graph networks (SGNs). Besides, there is also a method of using a typical detector to refine part features in~\cite{partreg,pgan,pmsm,aaver}. For example, part regularization \cite{partreg} uses you only look once (YOLO) \cite{yolo} as a detector to detect parts and feature extraction from part regions. Adaptive attention vehicle re-identification (AAVER)~\cite{aaver} uses a key-point detection module to localizing the part features and use an adaptive key-point selection module to learning the relationship of parts. Although part features have been widely used in supervised re-identification, unsupervised tasks have been challenging due to serious gradient collapse problems.

\section{Proposed Method}
%\vspace{-.2cm}
As shown in Fig. \ref{fig:framework}, the proposed Triplet Contrastive Representation Learning  (TCRL) framework consists of three components:
(1) a feature encoder module for extracting global and part features,
(2) a clustering module for generating pseudo labels,
and (3) three memory banks for storing updated features of the dataset, namely part memory bank, cluster memory bank, and instance memory bank.
Unlike other unsupervised re-identification methods, the input to feature encode module is two batches of images, i.e., the original and mask images.
Specifically, first, we sample a batch of original images and generate corresponding masked images.
And we use ResNet50~\cite{resnet} without a fully connected layer as the feature encode module.
Second, these two batches of images are fed to ResNet50 simultaneously to obtain global features and part features.
Third, a clustering algorithm (i.e., DBSCAN~\cite{dbscan}) is applied to cluster similar global features and assign pseudo labels to them.
The part feature of all samples, the average global feature of each cluster, and the global feature of all samples are stored in the part memory bank, cluster memory bank, and instance memory bank, respectively.
Finally, features of the three memory banks are updated with momentum via TCRL framework.
Moreover, TCRL  designs three different loss functions, i.e., Proxy Contrastive Loss~(PCL), Hybrid Contrastive Loss~(HCL), and Weighted Regularization Cluster Contrastive Loss~(WRCCL).
More detail about TCRL framework is described as follows.

\subsection{Preliminaries}
Assume that an unlabeled dataset $X = \{x_{1}, x_{2}, ..., x_{n}, ..., x_{N}\}$ consisting of $N$ original images.
For an original input image $x_{n} \in X $, correspondingly we generate a masked image ${x_{n}^{'}}$. Similarly, we can get the unlabeled masked dataset  $X^{'} = \{x_{1}^{'}, x_{2}^{'}, ..., x_{n}^{'}, ..., x_{N}^{'}\}$ of $N$ masked images.
We use both $x_{n} \in X$ and $x_{n}^{'} \in X^{'}$ as input images.
The global features ${F^{G}} = \{{f_{1}^{G}}, {f_{2}^{G}}, ..., {f_{n}^{G}}, ..., {f_{N}^{G}}\}$ and part features $F^{P} = \{f_{1}^{P}, f_{2}^{P}, ..., f_{n}^{P}, ..., f_{N}^{P}\}$ are obtained from the feature encode module.
To guide the contrastive learning, pseudo labels $Y_{K}$ are generated by global features through a clustering module. According to the pseudo labels, part memory bank $M^P$ and global memory bank $M^G$ are set as the current part features $F^{P}$ and global features $F^{G}$ before each forward propagation.

Different from the part memory bank $M^P$ and global memory bank $M^G$,  the mean global feature vectors of each pseudo labels are initialized with the
cluster memory bank $M^C = \{c_{1}, c_{2}, ..., c_{k}, ..., c_{K}\}$  by
\begin{equation}\label{eq:mc}
{c}_{k}=\frac{1}{\left|{M}^{C}_{k}\right|} \sum_{{f}_{i}^{G} \in {M}^{C}_{k}}{f}_{i}^{G},
\end{equation}
where ${M}^{C}_{k}$ represent the $k$-th cluster set of $M^C$ that contains all the feature vectors within cluster $k$ and $|\cdot|$ denotes the number of features in the set.
Noted that the clustering algorithm runs in each epoch , so the number of pseudo labels $K$ can be updated during the training phase.

\subsection{Proxy Contrastive Loss}

% 余弦loss + KLloss
A novel Proxy Contrastive Loss~(PCL) is proposed to indirectly model and transform instance-level (i.e., part and global) features via a cluster memory bank.
Note that this is not as easy as simply defining a loss function that includes both part and global branch. The reason is that they have different inputs and focus on different areas.
This choice is natural that PCL should contain two parts.
They make contrastive learning between adjacent memory banks, thus presenting the associations between the part and global features as a transition of the part-cluster and cluster-global associations.
Thus, the total of $L_{PCL}$ consists of two parts as follows,

\begin{equation}\label{eq:pcl}
L_{PCL}=\frac {L_{PCL}^{G} + L_{PCL}^{P}}{2},
\end{equation}
where the $L_{PCL}^{G}$ and $L_{PCL}^{P}$ denote the proxy contrastive learning loss of global and part features, respectively.

For simplicity we directly use $q_{i}$ and $q_{i}^{'}$ to represent the features of original-query image and masked-query image through the encoder module, except when specified.
Specifically, given the features of original-query image $q_{i}$ and the corresponding cluster feature $c_{k}$ from cluster memory bank $M^C$, we can defined the $L_{PCL}^{P}$ as follows,
\begin{equation} \label{eq:pclg}
L_{PCL}^{G}= L_{kl}\left( z(c_{k}), z(q_{i})\right)+ L_{dl}\left(q_{i}, c_{k}\right),
\end{equation}
where $L_{kl}$ is the kullback-leibler \cite{klloss} divergence loss, which enables the output logit value of query image $q_{i}$ to supervise the output logit value of cluster feature $c_{k}$;
$z(\cdot)$ denotes the softmax function.
$L_{dl}$ is the euclidean distance loss function to distill the relation between $q_{i}$ and $c_{k}$ by minimizing the distance.
 $L_{dl}$ is formulated as follows:
\begin{equation}\label{eq:dl}
L_{d l}\left(q_{i}, c_{k}\right)=\left\|q_{i} - c_{k}\right\|_{2},
\end{equation}
where ${||\cdot||}_{2}$  is $\ell_2$ normalization function.
Correspondingly, based on Eq. \eqref{eq:pclg} and Eq. \eqref{eq:dl}, assuming that features of masked-query image $q_{i}^{'}$, we can calculate $L_{PCL}^{P}$ of part feature as follows,
\begin{equation} \label{eq:pclp}
L_{PCL}^{P}= L_{kl}\left( z(c_{k}), z(q_{i}^{'})\right)+ L_{dl}\left(q_{i}^{'}, c_{k}\right).
\end{equation}
Since the cluster memory bank $M^C$ deals with all the features of vehicle, it can summarize them into a discriminative feature representation.

\subsection{Hybrid Contrastive Loss}
Given the feature of masked-query image $q_{i}^{'}$ along with pseudo label $y_k \in Y_{K}$, Hybrid Contrastive Loss (HCL) of the part feature ${L}_{HCL}^{P}$ is formulated as follows,

{\footnotesize
\begin{equation}\label{eq:hclp}
\begin{aligned}
L_{H C L}^P =
-\log \frac{\sum_{\substack{j=1 \\ j\in y_k}}^K \exp \left\langle q_i^{\prime} \cdot M_{j}^P / \tau\right\rangle}{\sum_{\substack{j=1 \\ j\in y_k}}^K \exp \left\langle q_i^{\prime} \cdot M_{j}^P / \tau\right\rangle+\sum_{\substack{n=1 \\ n \notin y_k}}^K \exp \left\langle q_i^{\prime} \cdot M_n^P / \tau\right\rangle},
\end{aligned}
\end{equation}}
where ${M}_{j}^{P}$ denotes the part features of positive instance with the same pseudo label as $q_{i}^{'}$. Instead,
${M}_{n}^{P}$ represents the part features of all negative samples from $M^P$, i.e., they do not belong to the same pseudo label $y_k$ as the current query sample $q_{i}^{'}$.
The $\tau$ is a temperature hyper-parameter, and set to 0.05.

In the same way, given the feature of original-query image $q_{i}$ along with pseudo label $y_k \in Y_{K}$, the HCL of global feature ${L}_{HCL}^{G}$ is defined as follows,
{\footnotesize
\begin{equation}\label{eq:hclg}
\begin{aligned}
L_{H C L}^P =
-\log \frac{\sum_{\substack{j=1 \\ j\in y_k}}^K \exp \left\langle q_i \cdot M_{j}^G / \tau\right\rangle}{\sum_{\substack{j=1 \\ j\in y_k}}^K \exp \left\langle q_i \cdot M_{j}^G / \tau\right\rangle+\sum_{\substack{n=1 \\ n \notin y_k}}^K \exp \left\langle q_i\cdot M_n^G / \tau\right\rangle},
\end{aligned}
\end{equation}}

 According to Eq. \eqref{eq:hclp} and Eq. \eqref{eq:hclg}, we calculate the distance between the query image and the feature vector of the instance.
 Ideally, HCL should be able to pull similar samples together rather than to use the cluster-level features (mean vectors of positive instance) in inter-class instances, like CCL~\cite{ccl}. The reason is that it is necessary to care for richer and differentiated positive sample information.
  Meanwhile, to add more negative sample information, we treat all samples except positive samples as negative samples instead of just using the mean vector of negative samples' clusters, such as SPCL~\cite{spcl}.
So our proposed HCL can close the distance between query samples and instance-level features of positive samples, pushing all negative samples away.

The two memory banks $M^{P}$ and $M^{G}$ are updated by using Eq. \eqref{eq:updatepg}, as follows,
\begin{equation}\label{eq:updatepg}
\begin{aligned}
{f}_{i}^{P}& \leftarrow \alpha {f}_{i}^{P}+(1-\alpha) {q}_{i}^{'} \\
{f}_{i}^{G}& \leftarrow \beta {f}_{i}^{G}+(1-\beta) {q}_{i},
\end{aligned}
\end{equation}
where $\alpha, \beta \in [0, 1]$ is a momentum constant used to control the update rate of memory banks.
$\alpha = \beta$ is set as 0.1.

\subsection{Weighted Regularization Cluster Contrastive Loss}

For pseudo labels $Y_{K}$, the results of clustering algorithms may be unreliable and bring noise samples.  We observe that images with correct labels are usually dominant, while images with wrong labels are from non-dominant uncertain classes. Therefore, we can judge whether the image belongs to a possible wrong label via measuring the similarity between the current query image and other images with the same pseudo label.

Formally, given the feature of query image $q_{i}$ along with pseudo label $y_k \in Y_{K}$, the weight $w_{i}$ is designed by using Eq. \eqref{eq:weight}, as follows,
\begin{equation}\label{eq:weight}
w_{i}=\frac{1}{N} \sum_{j=1}^{N} \frac{q_{i} \cdot q_{j}}{\left\|q_{i}\right\|_{2}\left\|q_{j}\right\|_{2}},
\end{equation}
where $N$ and ${||\cdot||}_{2}$ are denote the number with the same pseudo label $y_k$ as query image $q_{i}$ and the $\ell_2$ normalization function, respectively. Weighted Regularization Cluster Contrastive Loss (WRCCL) ${L}_\text{WRCCL}$ is further defined as:
\begin{equation} \label{eq:wrccl}
{L}_\text{WRCCL}=-w_{i}\log \frac{\exp <q_{i} \cdot c_{k}/ \tau>}{\sum_{j=1}^{K} \exp <q_{i} \cdot c_{j}/ \tau>},
\end{equation}
where $c_k$ represent the feature vector with the same pseudo label $y_k$ as the query image $q_{i}$ from cluster memory bank $M^C$.
According to Eq. \eqref{eq:wrccl}, we assign a lower weight to the training loss of the uncertain images in intra-class instances, so that the potentially correct images contribute more to cluster contrastive learning.
The cluster memory bank $M^C$  is updated according to Eq. \eqref{eq:updateccl}, as follows,

\begin{equation}\label{eq:updateccl}
c_{k} \leftarrow \gamma c_{k}+(1-\gamma){q_{i}},
\end{equation}
where $\gamma \in [0, 1]$ is a momentum constant same as $\alpha$ and $\beta$ in Eq. \eqref{eq:updatepg}.
Consistent with the update progress of memory banks $M^{P}$ and $M^{G}$, we set $\gamma = 0.1$ in the following experiments.

Thus, we propose a simple and unified TCRL framework that combining PCL, WRCCL, and HCL losses. The total loss function $L_{Total}$ of our proposed TCRL is as follows,
\begin{equation} \label{eq:overall}
L_{Total}= \lambda{(L_{PCL}^{P} + L_{PCL}^{G})} + \eta(L_{HCL}^{P} + L_{HCL}^{G}) +  {L_\text{WRCCL}},
\end{equation}
where $\lambda$ and $\eta $ are hype-parameters, used to control the balance between different losses.
Their default value are respectively set to 0.5 and 1.0 via cross-validation.
%\subsection{Loss Function}

\begin{table}[tp]
	\caption{The performance (\%) comparison on VeRi776. 'Source' denotes the source dataset. Best and second-best performance are in {\color{red}{red}} and {\color{blue}{blue}} color, respectively.} \label{tab:veri776}
%\vspace{-.2cm}
	\begin{center}
	%	\small
		%\setlength{\tabcolsep}{6pt}
\begin{threeparttable}
		\begin{tabular}{l|c||c|ccc}
			\hline
			\multicolumn{2}{c||} {Methods} &Source & mAP  &Rank1  &Rank5   \\
			\hline

\multirow{6}{*}{\begin{tabular}[c]{@{}c@{}}\textbf{Instance}\end{tabular}}
&MoCo \cite{moco}&None &9.53 &24.92 &40.61\\
&SimCLR \cite{simclr}&None &9.74 &25.42 &42.94\\
&SwAV  \cite{swav}&None &9.78 &25.86 &42.77\\
&BYOL \cite{byol}&None &9.92 &26.38 &44.68\\
&Simsiam  \cite{simsiam}&None &10.35 &28.84 &45.16\\
&Bottom \cite{bottom}& None&23.5 &63.7&73.4\\
\hline
\multirow{4}{*}{\begin{tabular}[c]{@{}c@{}}\textbf{Cluster}\end{tabular}}
&SPCL \cite{spcl}&VehicleID &38.9 &80.4 &86.8 \\
&CA-UReID \cite{ca-ureid}& None&40.08 &84.17 &88.25   \\
&CCL \cite{ccl} & None &40.3 &84.6 &{\color{blue}{89.2}} \\
&UCF \cite{ucf} &VehicleID &40.5 &{\color{blue}{85.2}} &88.7\\
\hline
\multirow{3}{*}{\begin{tabular}[c]{@{}c@{}}\textbf{Dual}\end{tabular}}
&MGCE-HCL \cite{mgce-hcl} &None &39.28 &81.56 &87.73 \\
&HHCL \cite{hhcl}&None &40.44 &85.33 &88.29 \\
&CACL \cite{cacl}&None &{\color{blue}{40.92}} &84.46 &88.21 \\
\hline
\multirow{7}{*}{\begin{tabular}[c]{@{}c@{}}\textbf{Others}\end{tabular}}
&OIM \cite{oim}& None&12.2 &45.1&62.2\\
&PUL \cite{pul} &None &17.06 &55.24 &66.27 \\
&HHL \cite{hhl} &None &17.52 &56.2 &67.61\\
&ECN \cite{ecn} &VehicleID &20.06 &57.41 &70.53\\
&DAN \cite{dan} &VehicleID &24.85 &58.46 &70.86\\
&UDAR \cite{udar} &VehicleID &35.8 &76.9 &85.8\\
&ML \cite{ml} &VehicleID &36.9 &77.8 &85.5\\
%&TDSR \cite{tdsr} &VehicleX &38.3 &85.1 &91.2\\
\hline
\multirow{1}{*}{\textbf{Proposed}}
&{TCRL}&None&{\color{red}{42.68}}&{\color{red}{87.26}} &{\color{red}{90.75}}\\
\hline
\end{tabular}
\end{threeparttable}
	\end{center}
\vspace{-0.4cm}
\end{table}

\section{Experiments and Analysis}

\subsection{Datasets}
\subsubsection{VeRi776}~\cite{veri776} is constructed by $20$ cameras in unconstrained traffic scenarios and each vehicle is captured by $2$-$18$ cameras.
Following the evaluation protocol of \cite{veri776}, VeRi776 is divided into a training subset containing $37,746$ images of $576$ subjects and a testing subset including a probe subset of $1,678$ images of $200$ subjects and a gallery subset of $11,579$ images of the same $200$ subjects.
% \vspace{-.2cm}
\subsubsection{VehicleID} \cite{drdl} totally includes $221,763$ images of $26,267$ subjects. The training subset consists of $110,178$ images of $13,164$ subjects. There are three testing subsets, i.e., Test800, Test1600, and Test2400, for evaluating the performance at different data scales.
Specifically, Test800 includes $800$ gallery images and $6,532$ probe images of $800$ subjects. Test1600 contains $1,600$ gallery images and $11,395$ probe images of $1,600$ subjects. Test2400 is composed of $2,400$ gallery images and $17,638$ probe images of $2,400$ subjects. Following the evaluation protocol of \cite{drdl}, for three testing subsets, the division of probe and gallery subsets is implemented as follows: randomly selecting one image of a subject to form the probe subset, and all remaining images of this subject are used to construct the gallery subset. This division is repeated and evaluated 10 times, and the average result is reported as the final performance.
% \vspace{-.2cm}

\begin{table*}[t]
	\caption{The performance (\%) comparison on VehicleID and Vehicle WILD. Best and second-best performance are in {\color{red}{red}} and {\color{blue}{blue}} color, respectively. Blanked
entries link to results not reported in previous works.}\label{tab:vehicleid}
% \vspace{-.4cm}
	\begin{center}
		%\normalsize
		\scriptsize
		\begin{tabular}{ll||ccccccc||ccccccc}
			\hline	
			\multicolumn{2}{c||}{\multirow{3}{*}{Methods}}
			&\multicolumn{7}{c||}{VehicleID}
            &\multicolumn{7}{c}{VERI-Wild}\\
            &
            &{Source}
			&\multicolumn{2}{c} {Test800}
			& \multicolumn{2}{c} {Test1600}
			& \multicolumn{2}{c||}{Test2400}
		    &{Source}
			&\multicolumn{2}{c} {Test3000}
			& \multicolumn{2}{c} {Test5000}
			& \multicolumn{2}{c}{Test10000}
			\\
			%            \cline{2-9}
%\multicolumn{2}{c||}{} & \multicolumn{1}{c}{}
& &
			 &mAP & Rank1
			 &mAP &Rank1
			&mAP &Rank1
			&  &mAP & Rank1
			 &mAP &Rank1
			&mAP &Rank1  \\
\hline
\multirow{2}{*}{\begin{tabular}[c]{@{}c@{}}\textbf{Instance}\end{tabular}}
&MoCo \cite{moco}&None &27.74 &22.68 &24.85 &19.51 &21.83 &15.80 &None &15.25 &38.70 &12.06 &34.81 &9.27 &31.22\\
&Simsiam  \cite{simsiam}&None &28.48 &23.21 &25.17 &19.94 &22.39 &16.55&None &15.66 &39.25 &12.43 &35.39 &9.65 &31.80\\
\hline

\multirow{3}{*}{\begin{tabular}[c]{@{}c@{}}\textbf{Cluster}\end{tabular}}
&SPCL \cite{spcl}&None &61.74 &55.46 &58.66 &51.58  &55.49 &47.92 &None &34.29 &71.38 &30.33 &64.82  &22.67 &61.13\\
&CA-UReID \cite{ca-ureid}& None&62.88 &56.50 &59.78 &52.32  &56.77 &49.05& None&36.18 &72.53 &31.44 &65.83  &23.92 &62.15 \\
&CCL \cite{ccl} & None &62.97 &56.71 &60.10 &52.55 &57.08 &49.33 & None &{\color{blue}{36.36}} &72.41 &31.42 &65.75  &23.86 &62.50 \\
\hline
\multirow{3}{*}{\begin{tabular}[c]{@{}c@{}}\textbf{Dual}\end{tabular}}
&MGCE-HCL \cite{mgce-hcl} &None&62.92 &56.69 &59.82 &52.78  &56.82 &49.08&None&35.77 &72.26 &31.15 &65.49  &23.57 &61.84\\
&HHCL \cite{hhcl}&None &63.60 &57.47 &60.95 &{\color{blue}{53.48}}  &57.04 &50.61&None &36.11 &72.60 &31.52 &65.76  &23.86 &62.44\\
&CACL \cite{cacl}&None &{\color{blue}{63.83}} &{\color{blue}{57.77}} &{\color{blue}{61.19}} &53.25  &{\color{blue}{57.47}} &{\color{blue}{50.80}} &None &36.28 &{\color{blue}{72.94}} &{\color{blue}{31.87}} &{\color{blue}{66.02}}  &{\color{blue}{24.33}} &{\color{blue}{62.99}}\\
\hline
\multirow{5}{*}{\begin{tabular}[c]{@{}c@{}}\textbf{Others}\end{tabular}}
&PUL \cite{pul} &None &43.90 &40.03 &37.68 &33.83 &34.71 &30.90&None&18.7&52.1&14.9&48.3&10.6&38.2\\
&DAN \cite{dan} &VeRi776 &49.53 &44.44 &43.90 &38.97 &40.07 &35.10&-&-&-&-&-&-&-\\
&ATTNet \cite{attnet} &VeRi776 &54.01 &49.48 &49.72 &45.18 &45.18 &40.71&-&-&-&-&-&-&-\\
&UDAR \cite{udar} &VeRi776 &59.6 &54.0 &55.3 &48.1 &52.9 &45.2  &None &30.0 &68.4 &26.2 &62.5 &20.8 &53.7\\
&ML \cite{ml} &VeRi776 &61.6 &54.8 &58.4 &51.3 &55.0 &47.5&-&-&-&-&-&-&-\\
&VAPC \cite{vapc} &-&-&-&-&-&-&- &None &33.0 &72.1 &28.1 &64.3 &22.6 &55.9\\
\hline
\multirow{1}{*}{\textbf{Proposed}}
&TCRL &None &{\color{red}{66.29}} &{\color{red}{60.36}} &{\color{red}{63.74}} &{\color{red}{56.22}} &{\color{red}{61.08}} &{\color{red}{52.93}}&None &{\color{red}{39.08}} &{\color{red}{75.22}} &{\color{red}{34.67}} &{\color{red}{68.59}} &{\color{red}{26.60}} &{\color{red}{64.31}}\\
			\hline
		\end{tabular}
	\end{center}
\vspace{-.4cm}
%\vspace{-.6cm}
\end{table*}

\subsubsection{VERI-Wild}~\cite{wild} has in total $416,314$ images of $40,671$ subjects divided into a training subset of $277,797$ images of $30,671$, and a testing subset of $128,517$ images of $10,000$ subjects.
Different to the VeRi776 \cite{veri776} and VehicleID \cite{drdl} captured at day, VERI-Wild also contains images captured at night.
Similar to VehicleID \cite{drdl}, the testing subset of VERI-Wild is organized into three different scale subsets, i.e.,
Test3000, Test5000, and Test10000.
Test3000 is composed of  $41,816$ gallery images and $3000$ probe images of $3,000$ subjects.
Test5000 is made up of  $69,389$ gallery images and $5,000$ probe images of $5,000$ subjects.
Test10000 is consisted of $138,517$ gallery images and $10,000$ probe images of $10,000$ subjects.
\subsection{Implementation Details} \label{ID}

Training configurations are summarized as follows.
(1) All the experiments are performed with 8 Nvidia Tesla V100 GPUs using the PyTorch \cite{pytorch} toolbox with FP16 training.
(2) We adopt ResNet50 \cite{resnet} as the backbone of the feature encoder and initialize the model with the parameters pre-trained on ImageNet.
(3) The input image is resized 224 $\times$ 224. Random horizontal flip and random crop are used for the data augmentation. Both probabilities of horizontal flip and crop are set to 0.5, respectively.
Noted that the occlusion area of the mask image generated by the original image is 0.2-0.4 times that of the original image, and the aspect ratio is 1.
(4) Each mini-batch includes 192 vehicle images, which includes 48 subjects and each subject holds $4$ images.
For the training phase, we use DBSCAN \cite{dbscan} for clustering to generate pseudo labels.
(5) The Adam optimizer  is applied to train parameters with  weight decays $5\times10^{-4}$.
There are $50$ epochs for the training process. The learning rates are initialized to $3\times10^{-4}$, and they are linearly warmed up to $3\times10^{-2}$ in the first $10$ epochs. After warming up, the learning rates are maintained at $3\times10^{-2}$ from $11$-th to $30$-th epochs.
Then, the learning rates are reduced to $3\times10^{-3}$ between $31$-th and $50$-th epochs.
Moreover, during the testing phase, the cosine distance of the global average pooling layer is applied as the similarity measurement for unsupervised vehicle re-identification.

\subsection{Performance Comparison}
For a clear presentation, we roughly divide the existing methods into four categories, namely ``Instance" \cite{moco,simclr,swav,byol,simsiam}, ``Cluster" \cite{spcl,ca-ureid,ccl,ucf}, ``Dual" \cite{mgce-hcl,hhcl,cacl}, and ``Others" \cite{oim,pul,hhl,bottom,ecn,dan,udar,ml} methods.

\begin{table}[tp]
	\caption{The TCRL ablation experiments on VeRi776. ``Directly" denotes directly models the part features and global
features. }\label{tab:framework_design}
	%\vspace{-.4cm}
	\begin{center}
		%\scriptsize

		\begin{tabular}{l||cccc }
			\hline
	
			{Setting} &mAP &Rank1 &Rank5 &Rank10

 \\
			\hline
   Directly  &2.19 &12.61 &19.84  &26.37 \\
   \hline
			Clustering in Part Branch			            &31.74 &70.28 &78.41  & 82.66     \\
			\hline
			w/o Part Branch			            &41.72 &85.17 &87.26  & 88.93     \\
\hline
           w/ Stop-gradient                        &42.47 &86.76   &89.61 & 90.85      \\
\hline
TCRL 			                            &\textbf{42.68} &\textbf{87.26}  &\textbf{90.75} &\textbf{91.68}       \\
			\hline
		\end{tabular}
	\end{center}
	%\vspace{-.4cm}
\end{table}

\subsubsection{Comparison on VeRi776}
From Table \ref{tab:veri776}, it can be found that the proposed TCRL method achieves the highest mAP (i.e., 42.68\%), rank1 (i.e., 87.26\%), and rank5 (i.e., 90.75\%), which respectively outperforms the CACL \cite{cacl} ($2$nd place) by 2.76\%, 2.80\%, and 2.54\%, due to considering part features, introducing all negative instances, and fixing mislabeled images.
We have also observed the ``Cluster", ``Dual", and ``Others" methods are mostly superior to the ``Instance" methods on the VeRi776 dataset by a large margin, indicating the importance of same-category correlations for unsupervised learning.
Then, compared to the ``Cluster" methods, the mAP and rank1 of the proposed TCRL approach exceeds 2.18\% and 2.06 \% over the best ``Cluster" method (i.e.,UCF \cite{ucf}).
It is noteworthy that UCF method uses an additional vehicle dataset (i.e., VehicleID), whereas our proposed TCRL does not require any additional training set.
Moreover, on mAP, rank1, and rank5, the proposed TCRL method is even significantly better than the ML \cite{ml} method using semantic information, which proves the direct introduction of part features can better learn fine-grained semantic information.

\begin{table*}[tp]
	\renewcommand{\arraystretch}{1.1}
	\caption{Evaluation the role (\%) of the different loss function on the VeRi776  and  VehicleID datasets. }\label{tab:ablation}
	%\vspace{-.6cm}
	\begin{center}
		%\scriptsize
		%\tiny
		%\setlength{\tabcolsep}{1pt}
		%\setlength{\tabcolsep}{1.2pt}
		\begin{tabular}{l||cc || cc cc cc}
			\hline
			%\noalign{\smallskip}
			\multirow{3}{*}{Setting}
			& \multicolumn{2}{c||}{VeRi776}&\multicolumn{6}{c}{VehicleID}\\
			& &
			& \multicolumn{2}{c}{Test800}
			& \multicolumn{2}{c}{Test1600}
			& \multicolumn{2}{c}{Test2400}
\\
			 &mAP& Rank1
			&mAP& Rank1
			&mAP& Rank1
			&mAP& Rank1
 \\

			\hline

			Baseline	($L_{CCL}$)			            &40.30 &84.60        &62.97 &56.71 &60.10 &52.55 &57.08 &49.33        \\
\hline
            { $L_{WRCCL}$} 	                &41.18 &85.52        &64.74 &58.68 &62.29 &54.07 &59.22 &51.55 \\
			{$L_{HCL}$} 		                &41.33 &85.79        &65.01 &58.93 &62.67 &54.42 &59.74 &51.96 \\
			{$L_{PCL}$}		                &41.58 &85.90        &65.26 &59.18 &62.84 &54.61 &59.79 &52.07        \\
\hline
            {$L_{ID}$ + $L_{Triplet}$} 		&38.82 &83.05       &59.27 &53.65 &55.18 &48.36 &53.40 &45.67\\
            {$L_{ID}$ + $L_{CCL}$} 		&40.38 &84.71        &63.12 &56.84 &60.25 &52.72 &57.23 &49.41\\
            {$L_{WRCCL}$ + $L_{HCL}$} 		&41.83 &86.64        &65.38 &59.41 &62.96 &54.88 &59.90 &52.31\\
			{$L_{WRCCL}$ +$L_{PCL}$} 		    &41.86 &86.59        &65.51 &59.67 &63.22 &55.06 &60.23 &52.44\\
			{$L_{HCL}$ + $L_{PCL}$}			&42.15 &86.47        &65.75 &59.92 &63.28 &55.27 &60.31 &52.47  \\
\hline
$L_{CCL}$ + $L_{ID}$ + $L_{Triplet}$   &40.61 &85.07         &63.33 &57.18  &60.49 &52.96 &57.46 &49.68   \\
TCRL ($L_{CCL}$ + $L_{ID}$ + $L_{Triplet}$) 	
   &41.34 &86.02         &64.74 &58.65  &61.88 &54.11 &59.23 &51.36   \\
\textbf{TCRL ($L_{WRCCL}$ + $L_{HCL}$ + $L_{PCL}$) }			
&\textbf{42.68}&\textbf{87.26}         &\textbf{66.29} &\textbf{60.36}  &\textbf{63.74} &\textbf{56.22} &\textbf{61.08} &\textbf{52.93} \\
			\hline
		\end{tabular}
	\end{center}
% 	\vspace{-.4cm}
\end{table*}

\subsubsection{Comparison on VehicleID}
In fact, the VehicleID \cite{drdl} dataset has a larger data scale than the VeRi776 \cite{veri776} dataset.
However, the proposed TCRL method still can obtain the $1$st place and outperforms those state-of-the-art methods under comparison, as occurred on the VeRi776 dataset, as shown in Table \ref{tab:vehicleid}.
For example, on Test800, Test1600, and Test2400, the proposed TCRL method respectively higher than the best ``Cluster" method, i.e., CCL \cite{ccl}, 3.65\%, 3.67\%, and 3.60\% on rank1.
Moreover, we compare our proposed TCRL with the most competing method CACL \cite{cacl}, which employs both instance memory bank and cluster memory bank for contrastive loss, but CACL underestimates part features and ignores all negative samples.
Based on the differences above, our TCRL method leads to 2.46\% improvements in mAP and up to 2.59\% gains in rank1 on Test800.

\subsubsection{Comparison on VERI-Wild}
The VERI-Wild \cite{wild} is a much larger dataset than VeRi776 \cite{veri776} and VehicleID \cite{drdl}, Table \ref{tab:vehicleid} shows that the proposed TCRL method wins the $1$st place among all compared state-of-the-art methods.
First, the ``Instance" methods (i.e., MoCo \cite{moco} and Simsiam \cite{simsiam}) can not acquire promising accuracies, which are inferior to the proposed
TCRL method and other three categories approaches.
Second, the proposed TCRL method has better performance than those ``Cluster" methods.
For example, taking the ``Cluster" methods with the cluster memory bank, i.e., CACL \cite{cacl}, it is still defeated by the proposed TCRL method, as it has lower mAP and rank1 on three different testing subsets (i.e., Test3000, Test5000 and Test1000).
Third, on largest Test1000 subset,mAP and rank1 of TCRL method respectively are 4.00\% and 8.41\% higher than those of the best ``Others" method, i.e., VAPC \cite{vapc}, which extra uses an viewpoint-aware module.
Meanwhile, the proposed TCRL method obtains the state-of-the-art performance on VeRi776, VehicleID, and VERI-Wild, which shows the effectiveness and robustness of our method.

\subsection{Ablation Studies}
In this section, we analyze the proposed TCRL from seven aspects:
(1) Advantage of TCRL design,
(2) Impact of each loss function,
(3) Influence of mask sampling strategy,
(4) Impact of  momentum value,
(5) Impact of  batch size,
(6) Role of different updating polices for cluster memory bank,
and (7) Qualitative Samples.

\begin{figure*}[tp]
	\centering
	\includegraphics[width=.9\linewidth]{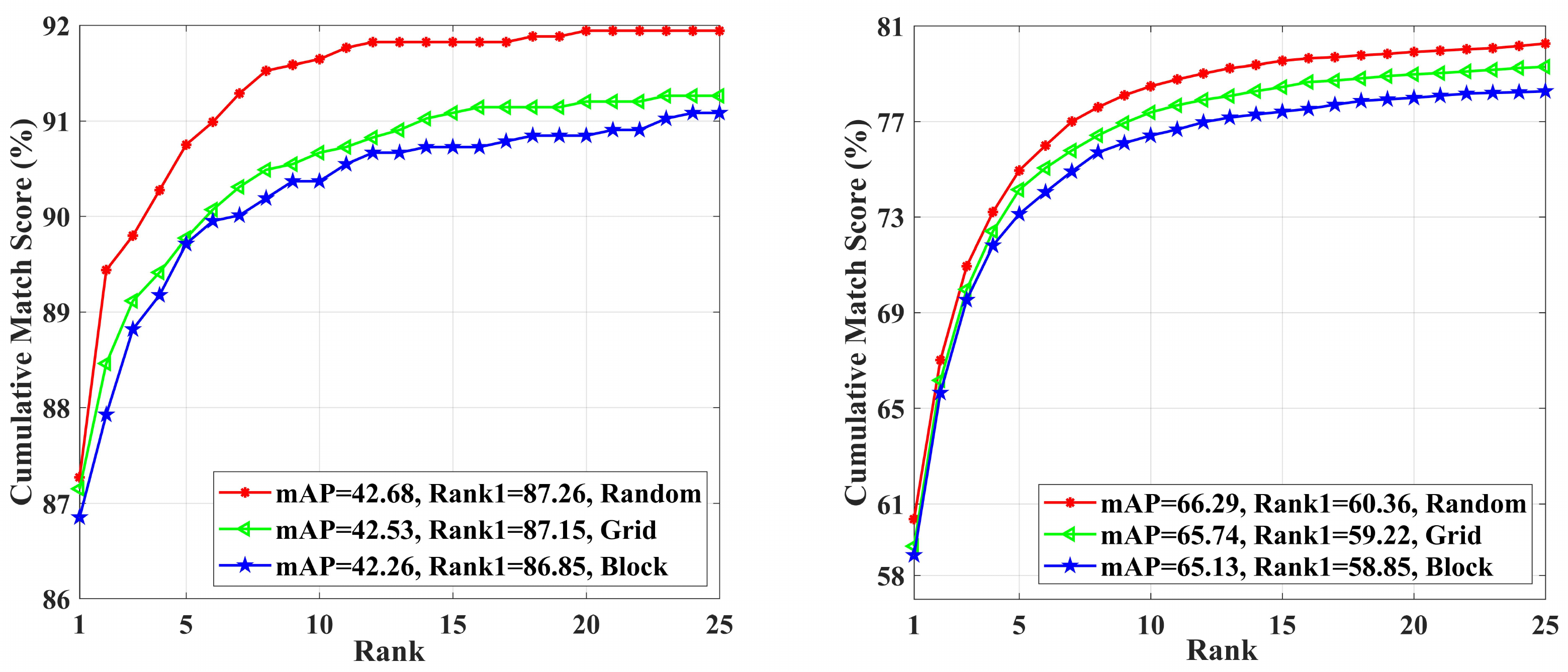}
		%\vspace{-.3cm}
	\caption{The CMC cures on VeRi776 and VehicleID. Different methods are compared from Rank1 to Rank25.}
	\label{fig:cmc}
	%	\vspace{-.4cm}
\end{figure*}

\begin{table*}[tp]
	\renewcommand{\arraystretch}{1.1}
	\caption{Different update policies for the cluster memory bank.}\label{tab:update}
	%\vspace{-.4cm}
	\begin{center}
		%\scriptsize
		%\tiny
		%\setlength{\tabcolsep}{1pt}
		%\setlength{\tabcolsep}{1.2pt}
		\begin{tabular}{l||cc || cc cc cc}
			\hline
			%\noalign{\smallskip}
			\multirow{3}{*}{Setting}
			& \multicolumn{2}{c||}{VeRi776}&\multicolumn{6}{c}{VehicleID}\\
			& &
			& \multicolumn{2}{c}{Test800}
			& \multicolumn{2}{c}{Test1600}
			& \multicolumn{2}{c}{Test2400}
\\
			 &mAP& Rank1
			&mAP& Rank1
			&mAP& Rank1
			&mAP& Rank1
 \\

			\hline
			Random				            &40.89 &85.17        &63.45 &57.83 &61.70 &54.03 &58.96 &50.51  \\
\hline
            Hard	                        &42.06 &86.76        &65.82 &59.77 &63.06 &55.85 &60.36 &51.93 \\
\hline
All 			                            &\textbf{42.68}&\textbf{87.26}         &\textbf{66.29} &\textbf{60.36}  &\textbf{63.74} &\textbf{56.22} &\textbf{61.08} &\textbf{52.93}   \\
			\hline
		\end{tabular}
	\end{center}
	%\vspace{-.4cm}
\end{table*}

\subsubsection{\textbf{Advantage of TCRL Design}}
To validate the effectiveness and superiority of TCRL, we pay special attention to designs that may affect model performance, whose result is shown in Table Table~\ref{tab:framework_design}.
The model does not work if it directly models the part features and global features. Collapsing is observed (first row of Table~\ref{tab:framework_design}) due to $M^P$ and $M^G$ are identity mapping. Then we tried to generate pseudo-labels using part features, but the performance (31.74\% vs. 42.68\% mAP) dropped significantly, indicating clustering based on part features is very difficult and not applicable.
We have also observed that a sufficient part feature is crucial role in vehicle re-identification. For example, TCRL respectively defeats w/o part branch 2.09\% and 3.46\% accuracy on Rank1 and Rank5.

Besides, we also find an interesting phenomenon, using stop-gradient is not necessary for TCRL design. As shown in Table~\ref{tab:framework_design}, w/ stop-gradient is comprehensively higher than clustering in part branch and w/o part branch in terms of mAP, Rank1, Rank5, and Rank10 accuracy, but only slightly lower than TCRL. Because we use a proxy strategy (i.e., cluster feature $c_k$) instead of directly performing contrastive learning with global and part features. Overall, it is reasonable and efficient for default TCRL  to give a solution on how to utilize part features in supervised re-identification.
\subsubsection{\textbf{Impact of Loss Functions}}
We conduct a set of experiments by disabling each loss function in our proposed TCRL individually, i.e., hybrid contrastive loss $L_{HCL}$, weighted regularization cluster contrastive loss $L_{WRCCL}$, and proxy contrastive loss $L_{PCL}$.
Noted that the 'Baseline' denotes the result of using only cluster contrastive loss (CCL) \cite{ccl}.
The ablation experimental results are shown in Table \ref{tab:ablation}.

From Table \ref{tab:ablation},  all setting methods have all consistently outperformed Baseline method on two datasets.
Especially $L_{WRCCL}$ is better than Baseline by more than 2.22 \% Rank1 on the largest Test2400, which demonstrates that $L_{WRCCL}$ can well penalize mislabeled images to improve performance.
Then, we can find that using both proposed loss functions simultaneously gives better results than using one loss function alone, which shows that different loss functions can be mutually compatible and mutually reinforcing. Furthermore, we find that the network using $L_{WRCCL}$, $L_{HCL}$,and $L_{PCL}$ instead of triplet loss or ID loss
could also reach a competitive performance.
For example, $L_{HCL}$ + $L_{PCL}$ respectively improves the performance of the $L_{ID}$ + $L_{Triplet}$ and $L_{ID}$ + $L_{CCL}$ by 6.91\% and 3.08\% mAP on the largest Test2400.

Besides, we show the performance drops when one loss function is disabled individually, as shown in Table \ref{tab:ablation}.
For example, TCRL respectively defeats the $L_{WRCCL}$ + $L_{HCL}$, $L_{WRCCL}$ +$L_{PCL}$, and $L_{HCL}$ + $L_{PCL}$  by 0.85 \%, 0.82\% and 0.53 \%in term of mAP on VeRi776.
These results show that each loss contributes to the performance improvements.
More importantly, compared to Baseline $(L_{CCL})$ and $L_{CCL}$ + $L_{ID}$ + $L_{Triplet}$, $TCRL(L_{WRCCL}$ + $L_{HCL}$ + $L_{PCL})$ achieves a total gain of 3.32\% and 2.96\% on mAP accuracy of Test800.
Because TCRL simultaneously constrains global, part,
and cluster features, CCL does not have a part branch
and only constrains the global features. This choice is natural since part features can boost performance further in supervised re-identification.

\subsubsection{\textbf{Influence of Mask Sampling}}
We compare three common mask sampling strategies, results as shown in Fig. \ref{fig:cmc}.
The Random denotes the default strategy used in this paper, see Section \ref{ID} for details.
The Grid represents that the default setting is used for grid mask, and the details of its can be found in \cite{gridmask}.
The Block means to randomly delete an area whose area is 30\% of the original image.
The 30\% is empirical results obtained through cross-validation.

Fig. \ref{fig:cmc} reveals an intuitive situation that from rank1 to rank25, the performance of Grid has consistently outperformed Block, which shows that the mask of the whole area (30\% entire image) is detrimental to learning discriminative features of vehicles.
Then, the Random method is significantly better than Grid and Block in the performance of rank1 and mAP on VeRi776 and VehicleID.
For example, Random method are 1.14\% and 1.51\% mAP higher than Grid and Block on Test800 of VehicleID.
Because Grid uses regular structured masks on all images, which easily leads the network to overfit this regular mask when learning part features.
In contrast, Random has a random irregular sampling with a high mask rate, which makes the network need to learn good representations for all the patches, and to mine discriminative part features from the patches.
These results demonstrate that simple random mask works best for our proposed TCRL method, resulting in good performance on two datasets.

\begin{figure*}[tp]
	\centering
	\includegraphics[width=1.0\linewidth]{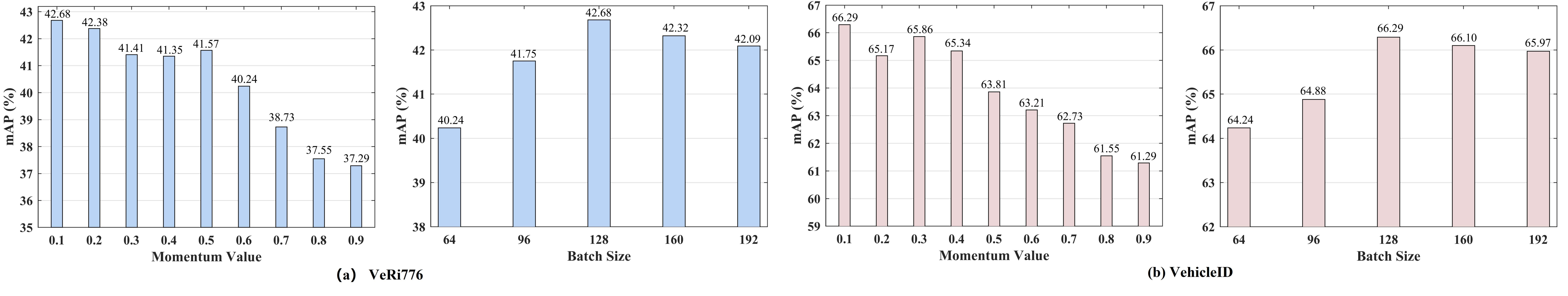}
		%\vspace{-.3cm}
	\caption{The impact of momentum value and batch size on VeRi776 and VehicleID. Different parameters are compared on mAP.}
	\label{fig:bs_value}
% 		\vspace{-.3cm}
\end{figure*}

\begin{figure}[tp]
	\centering
	\includegraphics[width=1.0\linewidth]{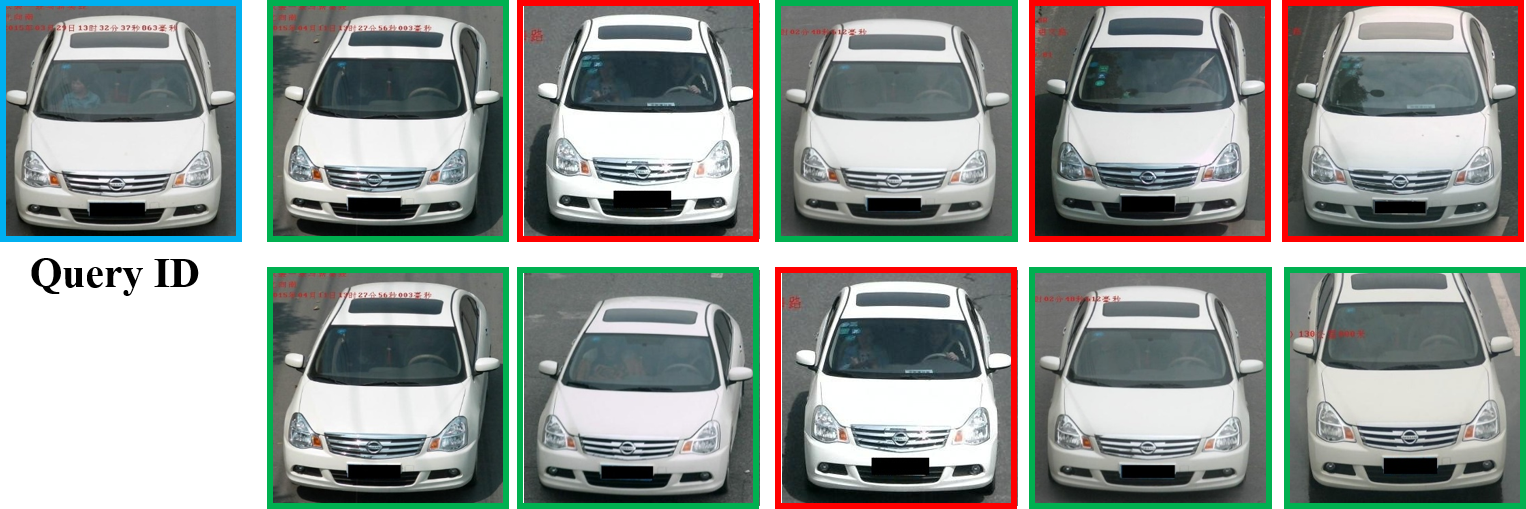}
	\caption{\footnotesize{Qualitative examples of cases. The first and second rows show the top five images returned by Baseline and TCRL, respectively. Images with blue, green, and red boxes denote query ID, correct, and incorrect retrieve results.}
	}
	\label{example}
\end{figure}

\subsubsection{\textbf{Impact of Momentum Value}}
As shown in Fig. \ref{fig:bs_value}, we adopt a momentum update strategy to refresh the part memory bank $M^P$, global memory bank $M^G$, and cluster memory bank $M^C$.
And the momentum value $\alpha$, $\beta$, and $\gamma$ controls the update speed of the memory banks.
The three memory banks use the same momentum value, that is, $\alpha=\beta=\gamma$ in the Eq. \ref{eq:updatepg} and Eq. \ref{eq:updateccl}.
From Fig. \ref{fig:bs_value}, when the momentum value is 0.1, the mAP performance is the highest on the VeRi776 and VehicleID datasets.
When the momentum value is greater than 0.5, the results of mAP drop significantly.
Therefore, we set $\alpha=\beta=\gamma=0.1$ in this paper.

\subsubsection{\textbf{Impact of Batch Size}}
We evaluate the performance impact of different batch sizes on proposed TCRL method.
The Fig. \ref{fig:bs_value}, shows the mAP performance for batch sizes from 64 to 192 on VeRi776 and VehicleID.
Overall, the performance of our method can remain stable in the batch size range of 64 to 192.
Compared with the state-of-the-art methods in Table \ref{tab:veri776} and Table \ref{tab:vehicleid}, our method achieves superior performance on regular batch sizes.
Especially, using a batch size of 128 can respectively get the highest 42.68\% and 66.29\% on two datasets, which is higher than the batch size of 64 and 192.
Thus we choose 128 as our default batch size setting.
%\subsubsection{Universality to Different Clustering Algorithms}

\subsubsection{\textbf{Role of Different Updating Polices for Cluster Memory Bank}}

There are three update strategies for cluster memory bank $M_C$, namely random update strategy, hard update strategy, and all update strategy.
The ``Random" and ``Hard" denote that we update the cluster memory bank $M_C$ with one random sample per class and the least similar sample in each class, respectively.
The ``All" indicates that all sample is used to update the cluster memory bank $M_C$.
The corresponding results are shown in Table \ref{tab:update}. The ``ALL" strategies achieve the highest 42.68\% mAP and 87.26\% rank1 on VeRi776.
Therefore, like most existing works~\cite{ccl}, we choose the 'All' update strategy for cluster memory bank $M_C$ in this work.

\subsubsection{\textbf{Qualitative Samples}}
To demonstrate some qualitative
results of our proposed TCRL, we present rank list visualization in Fig.~\ref{example}. Images with blue, green, and red boxes denote query ID, correct, and incorrect retrieve results, respectively.
The Rank1-5 errors of Baseline are often caused by vehicles with highly similar backgrounds and viewpoints, while the TCRL performed well and had more correct images in the rank list. Because we specially designed the $M^p$ and three different loss functions to focus on part features and penalize negative samples. These results demonstrate that the proposed TCRL can effectively capture the specific hints for each part.

\section{Conclusions}
This paper presents a simple Triplet Contrastive Representation Learning~(TCRL) framework, which leverages cluster features to bridge the part and global features.
Specifically, TCRL devises three memory banks to store the features according to their attributes. Then a Proxy Contrastive Loss~(PCL) is proposed to make contrastive learning between adjacent memory banks, thus presenting the associations between the part and global features as the part-cluster and the cluster-global associations.
To achieve higher performance, TCRL proposes two additional loss functions, the Hybrid Contrastive Loss~(HCL) to re-define the sample correlations by approaching the positive cluster features and leaving all the negative instance features, and the Weighted Regularization Cluster Contrastive Loss~(WRCCL) to refine the pseudo labels via penalizing the mislabeled images.
Extensive experimental results on three vehicle re-ID datasets, VeRi776, VehicleID, and VERI-Wild demonstrate that our method can be superior to state-of-the-art methods.
In the future, we hope our exploration will motivate people to rethink the roles of part features for unsupervised vehicle re-identification.

%\section*{Acknowledgment}


\ifCLASSOPTIONcaptionsoff
  \newpage
\fi



%\bibliographystyle{IEEEtran}
\bibliographystyle{IEEEtran}
\bibliography{reffullv2}



%\vfill


\end{document}